\documentclass[11pt]{article}

\usepackage{dmtw2015}
\usepackage{times}
\usepackage{url}
\usepackage{latexsym}

\usepackage{url}
\usepackage{graphicx}
\usepackage{float}
\usepackage{amsmath}

\usepackage{float}
\usepackage{textcomp}
\usepackage{listings}
\usepackage{multirow}
\usepackage{pbox}
\usepackage{lscape}
\usepackage{capt-of}

\usepackage{booktabs}
\usepackage{soul}
\usepackage[normalem]{ulem}
\usepackage{subfig}

\usepackage{tabularx,ragged2e}

\usepackage[noend]{algpseudocode}
\usepackage{caption}
\DeclareCaptionType{algorithm}

\usepackage{tikz}
\usepackage{pgfplots}
\usepackage{fixltx2e}

\usetikzlibrary{decorations.pathmorphing}

\newcolumntype{Y}{>{\centering\arraybackslash}X}

\PassOptionsToPackage{warn}{textcomp}

\newcommand{\tablehead}[1]{#1}
\newcommand{\specialcell}[2][l]{%
  \begin{tabular}[#1]{@{}l@{}}#2\end{tabular}}

\newcommand{\trankm}{t_\text{\scriptsize rank}}
\newcommand{\trank}{$\trankm$}

\newcommand{\tminsupportm}{t_\text{\scriptsize evd}}
\newcommand{\tminsupport}{$\tminsupportm$}

\newcommand{\tcosinem}{t_\text{\scriptsize sim}}
\newcommand{\tcosine}{$\tcosinem$}

\newcommand{\nPairsm}{N}
\newcommand{\nPairs}{$\nPairsm$}

\makeatletter
\def\blfootnote{\gdef\@thefnmark{}\@footnotetext}
\makeatother

\makeatletter
\newcommand\footnoteref[1]{\protected@xdef\@thefnmark{\ref{#1}}\@footnotemark}
\makeatother

\makeatletter
\newcommand\footnotesubref[2]{\protected@xdef\@thefnmark{\ref{#1}#2}\@footnotemark}
\makeatother

\newcommand{\afterfig}{\vspace{-0.35cm}}
\newcommand{\afterdoublefloat}{\vspace{-0.2cm}}

\newcommand{\dcar}{{\uparrow}d_{\mathrm{\emph{car}\underline{s}}}}
\newcommand{\walkvec}{{\uparrow}d_{\mathrm{\emph{walk}\underline{ing{\to}ed\vphantom{g}}}}}

\newcommand{\dfoot}{{\uparrow}d_{\mathrm{\underline{Haupt}\emph{ziel}}}}
\newcommand{\housevec}{{\uparrow}d_{\mathrm{\underline{dog}\emph{house}}}}

\title{Splitting Compounds by Semantic Analogy}

\newcommand{\illc}{\textsuperscript{$*$}}
\newcommand{\ai}{\textsuperscript{$\dagger$}}

\author{
{\centering\begin{tabular}[t]{c c c c} 
Joachim Daiber\illc & Lautaro Quiroz\ai & Roger Wechsler\ai & Stella Frank\illc \\[0.08in]
\multicolumn{2}{c}{\textnormal{\illc Institute for Logic, Language and Computation}} & \multicolumn{2}{c}{\textnormal{\ai Graduate School of Informatics}} \\ 
\multicolumn{2}{c}{\textnormal{University of Amsterdam}} & \multicolumn{2}{c}{\textnormal{University of Amsterdam}} \\
\multicolumn{2}{c}{\textnormal{Science Park 107, 1098 XG Amsterdam}} & \multicolumn{2}{c}{\textnormal{Science Park 904, 1098 XH Amsterdam}} \\[0.02in]
\multicolumn{2}{c}{{\tt \{J.Daiber,S.C.Frank\}@uva.nl}} & \multicolumn{2}{c}{{\tt first.last@student.uva.nl}}
\end{tabular}} 
}

\date{}

\begin{document}

\maketitle

\blfootnote{
\hspace{-0.65cm}
 This work is licenced under a Creative Commons 
 Attribution 4.0 International License.
}

\begin{abstract}
Compounding is a highly productive word-formation process in some languages
that is \mbox{often} problematic for natural language processing
applications.
%
In this paper, we investigate whether distributional semantics
in the form of word embeddings can enable a deeper,
i.e.,~more knowledge-rich, processing of compounds than the standard
string-based methods.
%
%
We present an unsupervised approach that exploits regularities in the
semantic vector space
(based on analogies such as ``bookshop is to shop as bookshelf is
to shelf'')
to produce compound analyses of high quality.
A subsequent compound splitting algorithm based on these analyses
is highly effective, particularly for ambiguous compounds.
German to English machine translation experiments show that this semantic analogy-based
compound splitter leads to better translations than a commonly used
frequency-based method.
\end{abstract}

\section{Introduction}
In languages such as German, compound words are a frequent occurrence leading to 
difficulties for natural language processing applications, and in particular machine translation.
Several methods for dealing with this issue---from shallow count-based methods
to deeper but more complex neural network-based processing methods---have been 
proposed.
The recent surge in practical models for distributional semantics has enabled
a multitude of practical applications in many areas, most recently in
morphological analysis \cite{soricut2015unsupervised}. 
In this paper, we investigate whether similar methods can be utilized to perform 
deeper, i.e.\ more knowledge-rich, processing of compounds.
A great asset of word embeddings are the regularities that their 
multi-dimensional vector space exhibits. 
\newcite{mikolov2013linguistic} showed that regularities such as 
``\textit{king} is to \textit{man} what \textit{queen} is to \textit{woman}''
can be expressed and exploited in the form of basic linear algebra operations
on the vectors produced by their method.
This often-cited example can be expressed as follows: 
$v(\mathrm{king}) - v(\mathrm{man}) + v(\mathrm{woman}) \approx v(\mathrm{queen})$,
where $v(.)$ maps a word into its word embedding in vector space.

In a very recent approach, \newcite{soricut2015unsupervised} exploit these 
regularities for unsupervised morphology induction. 
Their method induces vector representations for basic morphological transformations 
 in a fully unsupervised manner. String prefix and suffix replacement rules are induced directly from the data based on the idea that morphological processes can be modeled on the
basis of \textit{prototype} transformations, i.e.\ vectors that are good examples
of a morphological process are applied to a word vector to retrieve its inflected form.
A simple example of this idea is $\dcar = v(\mathrm{cars}) - v(\mathrm{car})$
and $v(\mathrm{dogs}) \approx v(\mathrm{dog}) + \dcar$, 
which expresses the assumption that the word \textit{car} is to \textit{cars} 
what \textit{dog} is to \textit{dogs}. 
The direction vector $\dcar$ represents the process of adding
the plural morpheme \emph{-s} to a noun.

While this intuition works well for frequently occurring inflectional morphology,
it is not clear whether it extends to more semantically motivated derivational processes 
such as compounding.
We study this question in the present paper.
Our experiments are based on the German language, in which compounding is a highly
productive phenomenon allowing for a potentially infinite number of
 combinations of words into compounds.
This fact, coupled with the issue that many compounds are observed infrequently
in data, leads to a data sparsity problem that hinders the processing 
of such languages.
Our contributions are as follows: 
After reviewing related work (Section~\ref{sec:relatedwork}), we study whether 
the regularities exhibited by the vector space also apply to
 compounds (Section~\ref{sec:motivation}).
We examine the relationship between the components
within compounds, as illustrated by the analogical relationship
``\emph{Hauptziel} is to \emph{Ziel} what 
\emph{Hauptader} is to
\emph{Ader}.''\footnote{In vector algebra: $\dfoot =
v(\mathrm{Hauptziel}) - v(\mathrm{Ziel})$ and $v(\mathrm{Hauptader})
\approx v(\mathrm{Ader}) + \dfoot$.
The compounds translate to main goal (\emph{Hauptziel}) and main
artery (\emph{Hauptader}). As a separate noun, \emph{Haupt} means
head.}
By leveraging this analogy we can then analyze the novel compound
\emph{Hauptmann} (captain) by searching for known string prefixes
(e.g.~\emph{Haupt-}) and testing whether the resulting split compound
(\emph{Haupt{\textbar}mann})
has a similar relation between its components
(\emph{haupt}, \emph{mann}) as the prototypical example
(\emph{Haupt{\textbar}ziel}).
We induce the compound components and their prototypes and apply them in a greedy compound splitting algorithm (Section~\ref{sec:method}), 
which we evaluate on a gold standard compound splitting task
(Section~\ref{par:evaluation}) and as a preprocessing step in a
machine translation setup (Section~\ref{sec:evaluation}).

\section{Related work}
\label{sec:relatedwork}
Our methodology follows from recent work on morphology induction
\cite{soricut2015unsupervised}, which combines string edits with distributional
semantics to split words into morphemes. In this model, morphemes are
represented as string edits plus vectors, and are linked into
derivation graphs. The authors consider prefix and 
suffix morphemes up to six characters in length; in contrast, our
approach to noun compound splitting only considers components at least
four characters long.

\subsection{Splitting compounds for SMT}

Dealing with word compounding in statistical machine translation (SMT) is essential to mitigate the
sparse data problems that productive word generation causes.
There are several issues that need to be addressed: \emph{splitting}
compound words into their correct components (i.e.\ disambiguating
between split points), \emph{deciding} whether to split a compound
word at all, and, if translating into a compounding language,
\emph{merging} components into a compound word (something we do not
address, but see \newcite{Fraser:2012} and \newcite{Cap:2014} for systems that do).
\newcite{koehn2003empirical} address German compound splitting using a
straightforward approach based on component frequency. They also
present splitting approaches based on word alignments and POS tag information, but find that while the more
resource-intensive approaches give better splitting performance 
(measured by gold-standard segmentations) the frequency-based method
results in the best SMT performance (measured by BLEU). This is
attributed to the fact that phrase-based MT system do not penalize
the frequency-based method for over-splitting, since it can handle
components as a phrase.


\newcite{Niessen:2000}, \newcite{Popovic:2006} and
\newcite{Fritzinger:2010} explore using morphological analyzers for
German compound splitting, with mixed results.
Since these approaches use heavy supervision
within the morphological analyzer, they are orthogonal to our
unsupervised approach.

It may be advantageous to split only
compositional compounds, and leave lexicalized compounds whole.
\newcite{Weller:2014} investigate this question by using distributional
similarity to split only words that pass a certain threshold (i.e.,
where the parts proposed by the morphological analyzer are similar to the compound).
Contrary to their hypothesis, they find no advantage in terms of SMT,
again indicating that oversplitting is not a problem for phrase-based
SMT.
The use of distributional similarity as a
cue for splitting is similar to the work presented in this paper.
However, the approach we follow in this paper is fully unsupervised,
requiring only word embeddings estimated from a monolingual corpus.
Additionally, it stands out for its simplicity, making it easy to
understand and implement.

\subsection{Semantic compositionality}

Noun compounding has also been treated within the field of
distributional semantics. \newcite{Reddy:2011} examine English noun
compounds
and find that distributional co-occurrence can capture the
relationship between compound parts and whole, as judged by humans in
terms of `literalness'.
\newcite{SchulteimWalde:2013} replicate this result for German, and also
show that simple window-based distributional vectors outperform
syntax-based vectors.

\section{Towards deeper processing of compound words}
\label{sec:motivation}

\subsection{Unsupervised morphology induction from word embeddings}

Our approach is based on the work of
\newcite{soricut2015unsupervised}, who exploit regularities in the
vector space to induce morphological transformations.
The authors extract morphological transformations in the form of
prefix and suffix replacement rules up to a maximum length of 6
characters.
The method requires an initial candidate set which contains all
possible prefix and suffix rules that occur in the monolingual corpus.
For English, the candidate set contains rules such as
\texttt{suffix:ed:ing}, which represents
the suffix \textit{ed} replaced by \textit{ing} (e.g.\
\textit{walked}$\to$\textit{walking}).
This candidate set also contains overgenerated rules that do not reflect
actual morphological transformations;
for example \texttt{prefix:S:$\epsilon$}\footnote{$\epsilon$ denotes
the empty string.} in~\textit{scream}$\to$\textit{cream}.

The goal is to filter the initial candidate set to remove
spurious rules while keeping useful rules.
For all word pairs a rule applies to, word embeddings are used to
calculate a vector representing the transformation.
For example, the direction vector for the rule \texttt{suffix:ing:ed}
based on the pair (\textit{walking}, \textit{walked})
would be $\walkvec = v(\mathit{walked}) - v(\mathit{walking})$.
%
For each rule there are thus potentially as many direction vectors as
word pairs it applies to.
A direction vector is considered to be meaning-preserving if it successfully
predicts the affix replacements of other, similar word pairs.
Specifically, each direction vector is applied to the first word in the
other pair and an ordered list of suggested words is produced.
For example, the direction vector $\walkvec$
can be evaluated against (\textit{playing}, \textit{played}) by
applying $\walkvec$ to \textit{playing} to produce the predicted
word form: $v(\mathit{played}*) = v(\mathit{playing}) + \walkvec$.
This prediction is then compared against the true word
embedding $v(\mathit{played})$ using a generic evaluation function
$E(v(\mathit{played}), v(\mathit{playing}) + \walkvec)$.\footnote{We
follow \newcite{soricut2015unsupervised} in defining $E$ as either
the \emph{cosine} distance or the \emph{rank} (position in the
predictions).} If the
evaluation function passes a certain threshold, we say that the
direction vector \emph{explains} the word pair.  Some direction
vectors explain many word pairs while others might explain very few.
To judge the explanatory power of a direction vector, a \emph{hit
rate} metric is calculated, expressing the percentage of applicable word pairs
for which the vector makes good predictions.\footnote{A transformation
is considered a \emph{hit} if the evaluated score is above a certain
threshold for each evaluation method $E$.}
Each direction vector has a hit rate and
a set of word pairs that it explains (its evidence set).
Apart from their varying explanatory power, morphological
transformation rules
are also possibly ambiguous. For example, the rule \texttt{suffix:$\epsilon$:s}
can describe both the pluralization of a noun (one
\emph{house}$\to$two \emph{houses}) and the 3rd person singular form
of a verb (I \emph{find}$\to$she \emph{finds}). Different
direction vectors might explain the nouns and verbs separately.

\newcite{soricut2015unsupervised} retain only the most explanatory
vectors by applying a recursive procedure to find the
minimal set of direction vectors explaining most word pairs.
We call this set of direction vectors \emph{prototypes},
as they represent a prototypical transformation for a rule and other words
are formed \emph{in analogy to} this particular word pair.
Finally, \newcite{soricut2015unsupervised} show that their prototypes can be 
applied successfully in a word similarity task for several languages.

\subsection{Compound words and the semantic vector space}

According to \newcite{lieber2009oxford}, compounds can be classified
into several groups based on whether the semantic head is part of the
compound (\emph{endocentric} compounds; a doghouse is a also a house) or
whether the semantic head is outside of the compound
(\emph{exocentric} compounds; a skinhead is not a head).
In this paper, we focus on endocentric compounds, which are also the
most frequent type in German.
Endocentric compounds consist of a modifier and a semantic head.
The semantic head specifies the basic meaning of the word and the
modifier restricts this meaning.
In German, the modifiers come before the semantic head; hence, the
semantic head is always the last component in the compound.
When applying the idea of modeling morphological processes by semantic analogy
to compounds, we can represent either the semantic head or the
modifier of the compound as the transformation (like the morpheme
rules above).
Since the head carries the compound's basic meaning, we
add the modifier's vector representation to the
head word in order to restrict its meaning.
We expect the resulting compound to be in the neighborhood of
the head word in the semantic space
(e.g., a \textit{doghouse} is close to \textit{house}).

\begin{figure}
\centering
\subfloat[Compounds with the same modifier.]{
  \includegraphics[height=4.4cm]{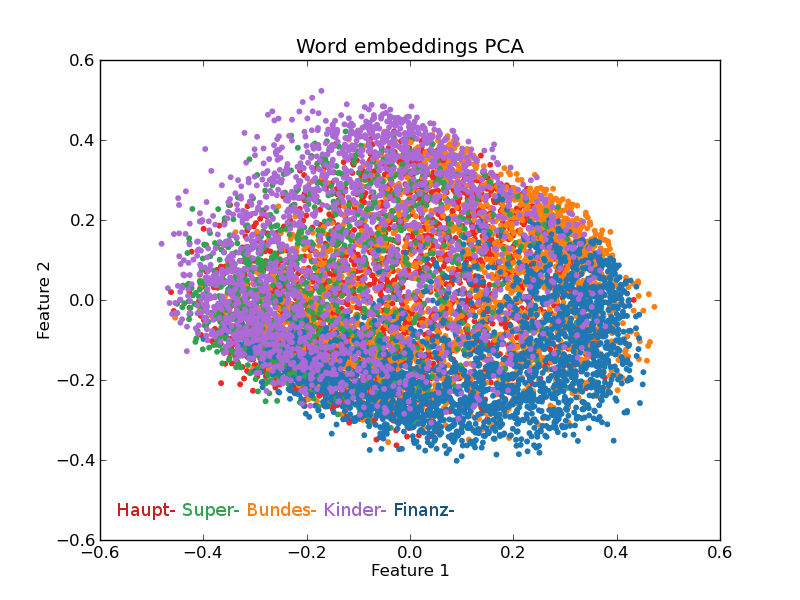}
  \label{fig:prefix_pca}
}
\subfloat[Compounds with the same head.]{
  \includegraphics[height=4.4cm]{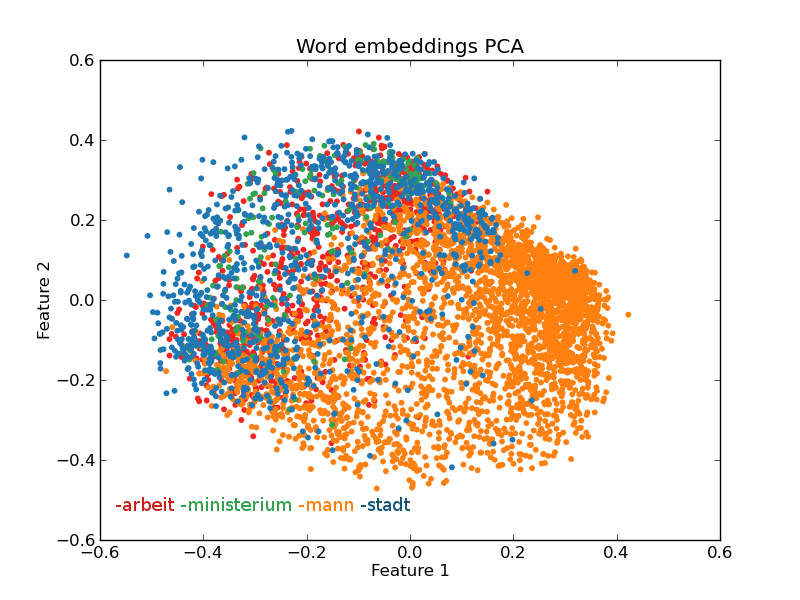}
  \label{fig:head_pca}
}
\afterdoublefloat
\caption{Semantic representations of compounds based on (a) their modifiers and (b) their heads.}
\label{fig:pca_plots}
\afterfig
\end{figure}

{
 \setcounter{footnote}{\value{footnote}+1}
 \footnotetext{\label{gloss1}Gloss for modifiers: (a) main, (b) federal, (c) children, (d) finance. Heads: (e) piece of work, (f) ministry, (g) man, (h) city.}
}
We illustrate this intuition by visualizing compound words and their
parts in the vector space.
All visualizations are produced by performing principal component analysis (PCA)
to reduce the vector space from 500 to 2 dimensions.
Figure~\ref{fig:pca_plots} presents the visualization of various compounds with either
the same head or the same modifier.
For Figure~\ref{fig:prefix_pca}, we plot all German compounds in our dataset 
that have one of the modifiers \textit{Haupt-,\footnotesubref{gloss1}{a} Super-, Bundes-,\footnotesubref{gloss1}{b} Kinder-\footnotesubref{gloss1}{c}} or \textit{Finanz-}.\footnotesubref{gloss1}{d} 
Figure~\ref{fig:head_pca}, on the other hand, shows a plot for all 
German compounds that have one of the heads \textit{-arbeit,\footnotesubref{gloss1}{e} -ministerium,\footnotesubref{gloss1}{f} \mbox{-mann}\footnotesubref{gloss1}{g}} 
or \mbox{\textit{-stadt}}.\footnotesubref{gloss1}{h}
Hence, the two plots illustrate the difference between learning vector
representations for compound modifiers or heads.
Words with the same modifier do not necessarily appear in close proximity 
in the embedding space.
This is particularly true for modifiers that can be applied liberally to 
many head words, such as \textit{Super-} or \textit{Kinder-}.\footnotesubref{gloss1}{c}
On the other hand, compounds with the same head are close 
in the embedding space. 
This observation is crucial to our method, as we aim to find
direction vectors that generalize to as many word pairs as possible.


\section{Compound induction from word embeddings}
\label{sec:methodology}
\label{sec:method}

\subsection{Compound extraction}
\label{sec:compound_extractation}

\paragraph{Candidate extraction}
\label{sub:candidate_extraction}

We compile an initial set of modifier candidates by extracting all possible 
prefixes with a minimum length of 4 characters.\footnote{For efficient 
computation, we use a directed acyclic word graph: \url{https://pypi.python.org/pypi/pyDAWG}.}
We retain a modifier as a candidate if both the modifier and the rest of the word, 
i.e.\ the potential head of the compound, occur in the vocabulary. 
The initial candidate set contains 281K modifiers, which are reduced 
to 165K candidates by removing the modifiers occurring in only one word.
The length of the average support set (i.e., the set of all compounds the modifier applies to) is 13.5 words.
Table~\ref{tbl:candidates_example} shows the ten candidate modifiers with the 
biggest support sets. 
At this stage, the candidate set contains any modifier-head split that can be 
observed in the data, including candidates that do not reflect real 
compound splits.\footnote{For example, as \textit{Para} (a river) and \textit{dies} (this) 
occur in the data, an incorrect candidate split occurs for \textit{Para{\textbar}dies} (paradise).}
Compound splits are not applied recursively here, as we assume
that internal splits can be learned from the occurrences of the heads as individual words.\footnote{For example, for \emph{Haupt{\textbar}bahn{\textbar}hof} (main train station), we observe both \emph{Haupt{\textbar}bahnhof} and \emph{Bahn{\textbar}hof}.
}

\paragraph{Prototype extraction}
\label{sub:prototype_extraction}

To find the prototype vectors that generalize best over the most words in the support set, 
we apply the same recursive algorithm as \newcite{soricut2015unsupervised}.
The algorithm initially computes the direction vector for each 
\emph{(modifier, compound)} pair in the support set by subtracting the embedding
of the head from the embedding of the compound, e.g.\ $\housevec =
v(\mathit{doghouse}) - v(\mathit{house})$. 
Each direction vector is then evaluated by applying it to all the word pairs 
in the support set, for example ${v(\mathit{owner}) + \housevec \stackrel{?}{=}
  v(\mathit{dogowner})}$ 
for the word pair \textit{dog{\textbar}owner}.
If the resulting vector is close (according to $E$) to 
the vector of the actual target compound, we add it to the evidence set 
of the vector.
The direction vector with the largest evidence set is selected as a
prototype.
All pairs this prototype explains are then removed and the algorithm is applied recursively
until no direction vector explains at least \tminsupport{} compounds.
As the evaluation function $E$ we use the rank of the
correct word in the list of predictions and experiment with
$\tminsupportm = \{ 10, 6, 4 \}$.
Lastly, for efficient computation we sample the evidence set down to a maximum 
number of $500$ words.

\addtocounter{footnote}{1}
\footnotetext{\label{gloss2}Words are related to mouse pointer (\emph{Zeiger}), biological genus (\emph{St\"amme}), mouse costume (\emph{Kost\"um}) and control (\emph{Steuerung}).}
\newcommand{\z}{\phantom{0}}
\begin{table}[t]
\centering
\small
\subfloat[Modifiers by size of support set.]{%
\begin{tabular}{llc l llc}
\toprule
& \tablehead{Modifier} & \tablehead{Support}&  & & \tablehead{Modifier} & \tablehead{Support}\\
\toprule
1. & {\it Land-}   & 8387 &&\z6.&  {\it Landes-} & 5189                    \\
2. & {\it Kinder-} & 6249 &&\z7.&  {\it Schul-}  & 5011                    \\
3. & {\it Haupt-}  & 5855 &&\z8.&  {\it Jugend-} & 4855                    \\
4. & {\it Lande-}  & 5637 &&\z9.&  {\it Ober-}   & 4799                    \\
5. & {\it Stadt-}  & 5327 && 10.&  {\it Gro\ss-} & 4656                    \\
\bottomrule

\end{tabular}
\label{tbl:candidates_example}
}\hspace{0.5cm}
\subfloat[Prototypes and evidence words for \emph{Maus-}.\protect\footnoteref{gloss2}]{%
\begin{tabular}{l l}
\toprule
\tablehead{Prototype} & \tablehead{Evidence words}      \\
\toprule
$v_\text{-Zeiger}$          & \specialcell[t]{-Bewegung -Klicks -Klick -Tasten\\ -Zeiger} \\
$v_\text{-St\"amme}$        & \specialcell[t]{-Mutanten -Gene -Hirnen -St\"amme}         \\
$v_\text{-Kost\"um}$        & -Knopf -Hirn -Hirns -Kost\"um             \\
$v_\text{-Steuerung}$       & -Ersatz -Bedienung -Steuerung           \\
\bottomrule
\end{tabular}
\label{tbl:mouse_example}
}
\afterdoublefloat
\caption{Overall most common modifiers and the prototypes extracted for the modifier \emph{Maus-}.}
\afterfig
\end{table}

\subsection{Implementation considerations}
\label{sub:implementation_considerations}

We now turn to implementation considerations and perform an intrinsic
evaluation of the prototypes.

\paragraph{Word embeddings} We use the German data of the \textit{News Crawl Corpora} (2007-2014).\footnote{\label{wmtnote}\url{http://www.statmt.org/wmt15/translation-task.html}} 
The text is truecased and tokenized, and
all punctuation characters are removed, resulting in approximately 2B
tokens and a vocabulary size of 3M.
We use \textit{word2vec} to estimate the word embeddings.\footnote{\url{https://code.google.com/p/word2vec/}}
We train 500-dimensional word embeddings using the skip-gram model, a window 
size of 5 and a minimum word frequency threshold of 2. 
The latter ensures that we find word embeddings for all words that occur at 
least twice in the corpus, which is useful as long compounds may occur only
very few times.

\paragraph{Treatment of interfixes (Fugenelemente)}
\label{par:treatment_of_interfixes}

For mostly phonetic reasons, German allows the insertion of a limited set of 
characters between the modifier and the head.
As learning this set is not the aim of our work, we
simply allow the fixed set of interfixes $\{$\textit{-s-}, \textit{-es-}$\}$ to
occur.
For any combination of interfix and casing of the head word, we add the tuple 
of the two to the support set of the corresponding modifier.

\paragraph{What do the prototypes encode?}
\label{par:what_do_the_prototypes_encode_}

An inspection of the prototypes for each modifier shows that the differences between
them are not always clear cut. 
Often, however, each prototype expresses one specific sense of the modifier.
Table \ref{tbl:mouse_example} illustrates this on the example of the German 
modifier \textit{Maus-} (Engl.\ \textit{mouse}), which can refer to both 
the animal and the computer device. 
Although there are more than two prototype vectors, it is interesting to 
observe that the two word senses are almost fully separated.



\paragraph{Calculating the hit rate}
\label{par:calculating_the_hit_rate}

To evaluate the quality of the prototypes, we use the hit rate
metric defined by \newcite{soricut2015unsupervised}.
A direction vector's hit rate is the percentage of relevant word pairs that can be explained by the vector.
A prediction is explainable if the actual target word is among the top
\trank~predictions
and, optionally, if the cosine similarity between the two is at least~\tcosine.

The implementation of this evaluation function $E$ requires the calculation
of the cosine distance between a newly created vector and the word vector
of every item in the vocabulary.
Since this score is calculated \nPairs~times for every
of the \nPairs~word pairs (i.e., $\nPairsm^2$ times), this is a computationally 
extremely expensive process.
For more efficient computation, we use an approximate k-nearest neighbor 
search method.\footnote{\url{https://github.com/spotify/annoy}}
While this is not a lossless search method, it offers an adjustable trade-off 
between the model's prediction accuracy and running time.\footnote{With this fast approx.\ search method the total training time would be just below 7 days if run on a single 16 core machine.} 
For a standard setting ($\tminsupportm=6,\trankm=80$), the hit rates using
approximate and exact rank are 85.9\% and 60.9\% respectively.
This shows that the hit rates obtained with the approximate method are 
more optimistic, which will affect how the prototype vectors are extracted.
Additionally, restricting both \emph{rank} and \emph{similarity} ($\trankm=80, \tcosinem \geq 0.5$)
leads to lower hit rates (25.9\% for approximate and 15\% for exact rank).

\paragraph{Influence of thresholds}
\label{par:influence_of_thresholds}
Table~\ref{tbl:prototypes_results} compares the parameters of our
model based on (a) the mean hit rate, (b) cosine similarity, (c) the percentage
of candidate modifiers with at least one prototype and (d) the mean number of
prototypes per rule.
Higher values of \tminsupport~(minimum evidence set size) lead to better quality in terms
of hit rate and cosine similarity as prototypes have to be able to cover a 
larger number of word pairs in order to be retained.
The rank threshold \trank{} also behaves as expected. 
Reducing \trank~to 80 means that the predicted vectors are of higher quality 
as they need to be closer to the true compound embeddings. 
Tables~(c) and (d) illustrate that the more restrictive parameter settings 
reduce the amount of modifiers for which prototypes can be extracted. 
From a total of 165399 candidate prefixes, only 3\%-10\% are retained in 
the end for our settings. Similarly, the average number of prototypes per 
modifier also decreases with more restrictive settings. 
Interestingly, however, for the most restrictive setting 
($\tminsupportm=10$, $\trankm=80$), this number is still a relatively high 2 
prototypes per vector.

\begin{table}[t]
  \small
  \centering
  \begin{tabularx}{\textwidth}{@{} l@{} YY c YY c YY c YY @{}}
  \toprule
                         & \multicolumn{2}{@{}c@{}}{(a) Mean hit rate} & & \multicolumn{2}{@{}c@{}}{(b) Mean cosine sim.} & & \multicolumn{2}{@{}c@{}}{(c) \% with prototypes} & & \multicolumn{2}{@{}c@{}}{(d) Mean \# of prototypes} \\
                           \cmidrule{2-3} \cmidrule{5-6} \cmidrule{8-9} \cmidrule{11-12} 
  \hspace{1.5cm}$\trankm=$ & $80$ & $100$ & & $80$ & $100$ &  & $80$ & $100$ & & $80$  & $100$ \\
  \toprule 
  $\tminsupportm = 4 $   & 26\%  & 22\%   & & 0.39  & 0.39   &  & 8.93\% & 9.52\%  & & 4.20  & 4.16  \\
  $\tminsupportm = 6 $   & 31\%  & 26\%   & & 0.43  & 0.43   &  & 5.13\% & 5.47\%  & & 3.29  & 3.30  \\
  $\tminsupportm = 10$   & 36\%  & 31\%   & & 0.45  & 0.45   &  & 2.91\% & 3.14\%  & & 2.25  & 2.29  \\
  \bottomrule
  \end{tabularx}
  \caption{Overview of the influence of the hyperparameters on prototype extraction.}
  \label{tbl:prototypes_results}
  \afterfig
\end{table}

\subsection{Compound splitting}
\label{sub:compound_splitting}

To obtain a clearer view of the quality of the extracted compound
representations, we apply the prototypes to a compound splitting task.

\paragraph{Splitting compounds by semantic analogy} 
\label{par:splitting_compounds_by_analogy}

\newcommand{\algS}[1]{\mathit{#1}}
\newcommand{\algWord}{\algS{word}}
\newcommand{\algV}{\algS{V}}
\newcommand{\algM}{\algS{M}}

\newcommand{\algModifier}{\algS{modifier}}
\newcommand{\algModifiers}{\algS{modifiers}}

\newcommand{\algBestModifier}{\algS{bestModifier}}

\newcommand{\algHead}{\algS{head}}
\newcommand{\algProtoHead}{\algS{head}_\algS{proto}}
\newcommand{\algProtoWord}{\algS{word}_\algS{proto}}

The extracted compound modifiers and their prototypes can be employed 
directly to split a compound into its components.
Algorithm~\ref{alg:compoundsplit} presents the greedy algorithm applied to every
word in the text. $\algV$ is the word embedding vocabulary, $\algM$ is the set of extracted
modifiers with their prototypes, and \textsc{Prefixes}(.) is a function returning 
all string prefixes.

\begin{algorithm}[h]
\begin{algorithmic}[1]
\Procedure{Decompound}{$\algWord$, $\algV$, $\algM$}
  \State $\algModifiers \gets \left\{ m \mid p \gets \Call{Prefixes}{\algWord} \text{ if } p \in \algM \right\}$
  \If {$\algModifiers = \emptyset$ OR $\algWord \notin \algV$} \label{algl:conditions1}
    \State \textbf{return} $\algWord$
  \EndIf
  \State $\algBestModifier \gets \emptyset$
  \For{$\algModifier \in \algModifiers$}
    \State $\algHead \gets \algWord$ without $\algModifier$ \Comment{e.g.\ house $\gets$ doghouse without dog-}
    \If {$\algHead \in \algV$} \label{algl:conditions2}
      \For{$(\algProtoHead, \algProtoWord) \in \algModifier$}
        \State Evaluate ``$\algWord$ is to $\algHead$ what $\algProtoWord$ is to $\algProtoHead$''\label{algs:eval}
        \\\Comment{e.g.\ doghouse is to house what dogowner is to owner}
        \State Update $\algBestModifier$ if this is the best match so far
      \EndFor
    \EndIf
  \EndFor
  \State \Return $\algWord$ split based on $\algBestModifier$
\EndProcedure
 \end{algorithmic}
\afterdoublefloat
\caption{Greedy compound splitting algorithm.}
\label{alg:compoundsplit}
\afterfig
\end{algorithm}

Compounds may only be split if 
(a) the full compound word is in the vocabulary $\algV$, i.e.\ it has been observed at least twice in the training data (Line~\ref{algl:conditions1}),
(b) it has a string prefix in the modifier set and this modifier has at least one prototype (Line~\ref{algl:conditions1}), 
(c) the potential head word resulting from splitting the compound based on the modifier is also in our vocabulary (Line~\ref{algl:conditions2}).
The last case, namely that the compound head candidate is not in the vocabulary 
can occur for two reasons: either this potential head is a valid word that has not
been observed frequently enough or, the more common reason, the substring is not a 
valid word in the language.\footnote{For example, when applying the algorithm to \emph{Herrengarderobe} (male cloak room), two possible prefixes apply: \emph{Herr} and \emph{Herren}. In the first case, the remaining slice is \emph{engarderobe}, which is not a valid word and thus the candidate prefix is discarded.}
The algorithm's coverage can be increased by backing off to a frequency-based
 method if conditions (a) or (c) are violated. 
The core of the algorithm is the evaluation of meaning preservation in 
Line~\ref{algs:eval}.
This evaluation is performed using the \emph{rank}-based and \emph{cosine similarity}-based
evaluation functions. 
Modifiers that do not pass the thresholds defined for these functions
are discarded as weak splits.
To split compounds with more than two components, the algorithm is applied
recursively.

\paragraph{General evaluation}
\label{par:evaluation}

We use the test set from \newcite{henrich2011determining}, which contains a list
 of 54569 compounds annotated with binary splits. 
As we only consider prefixes with a minimal length of 4 characters, we 
filter the test set accordingly, leaving 50651 compounds.
Moses \cite{koehn2007moses} offers a compound splitter that splits a word if the geometric average of the frequencies of its components is higher than the frequency of the compound.
We trained two instances of this compound splitter to use as references: 
one using the German monolingual dataset used to train the Word2Vec models and a second using a subset of the previous dataset.\footnote{Subset: \emph{News Crawl 2007-2009} (275M tokens, 2.09M types). Full set: \emph{News Crawl 2007-2014} (2B tokens, 3M types).}
Unlike our method, the two baseline systems do not consider the meaning 
preservation criteria of the compound splitting rules that are applied.
Results for the full test set (accuracy and coverage, i.e.\ $\frac{|\text{correct splits}|}{|\text{compounds}|}$ and $\frac{|\text{compounds split}|}{|\text{compounds}|}$) are presented in the first row of Table~\ref{fig:amb_eval}.

\paragraph{Evaluation of highly ambiguous compounds}

\newcommand{\potentialsplit}{\hspace{1pt}\begin{tikzpicture}[baseline,every node/.style={anchor=base}]
  \draw[decorate,decoration={snake, segment length=0.7mm, amplitude=0.1mm}] (0,-0.03) -- (0,0.23);
\end{tikzpicture}\hspace{1pt}}

The strength of our method resides in the capacity to discriminate good 
candidate splits from bad ones.
By capturing the meaning relation between compounds and their
components, we are able to decide for a given word which splitting
rule is the most appropriate.
With this in mind, our approach should stand out in contexts where
multiple split points may apply to a compound.
We simulate different ambiguity scenarios based on 
Henrich and Hinrich's gold standard dataset:
We extract compounds for which we find 
2, 3, 4, and 5 potential split points.\footnote{Each string prefix which occurs as a separate word produces a potential split point (indicated by \potentialsplit).
The potential split points may not be linguistically motivated and can lead to correct (\emph{general\textbar{}stabs}) or incorrect splits (\emph{gene\potentialsplit{}rals\potentialsplit{}tabs}).
Examples include 
Einkauf{\potentialsplit}s{\potentialsplit}wagen, Eis{\potentialsplit}en{\potentialsplit}bahn{\potentialsplit}unternehmen, Wissen{\potentialsplit}s{\potentialsplit}chaft{\potentialsplit}s{\potentialsplit}park and Gene{\potentialsplit}ra{\potentialsplit}l{\potentialsplit}s{\potentialsplit}tab{\potentialsplit}s.
}
The resulting test sets consists of 18571, 1815, 842 and 104 compounds, respectively.
For all compound splitting experiments, we use the prototype vectors 
extracted with the parameters $\tminsupportm=6$ and $\trankm=100$.

\newcommand{\acc}{Acc.}
\newcommand{\cov}{Cov.}

\definecolor{burgundy}{rgb}{0.5, 0.0, 0.13}
\definecolor{navyblue}{rgb}{0.0, 0.0, 0.5}
\definecolor{darkolivegreen}{rgb}{0.33, 0.42, 0.18}
\definecolor{dartmouthgreen}{rgb}{0.0, 0.5, 0.0}

\newsavebox{\tempbox}
\begin{table}
\sbox{\tempbox}{
\begin{tikzpicture}[thick,scale=0.55, every node/.style={scale=0.65}]
    \begin{axis}[
        name=ambplot,
        xlabel=\textsc{Ambiguity},
        ylabel=Accuracy and Coverage (\%),
        width=10cm,height=6.4cm,
        legend style={font=\Large,
                    at={(0.5,-0.16)},
                    anchor=north,
                    legend columns=3,
                        },
        minor x tick num=0, xtick={2,3,4,5},
    ]


    \addplot[color=burgundy,style=dotted,mark=diamond*,mark options={scale=1.1,solid}] plot coordinates {
        (2, 24.94)
        (3, 21.1 )
        (4, 22.09)
        (5, 24.04)        
    };

    \addplot[color=navyblue,style=dotted,mark=square*,mark options={scale=1.1,solid}] plot coordinates {
        (2, 13.13)
        (3, 8.04 )
        (4, 9.98 )
        (5, 9.62 )        
    };

    \addplot[color=dartmouthgreen,style=dotted,mark=*,mark options={scale=1.1,solid}] plot coordinates {
        (2, 1.79)
        (3, 1.21)
        (4, 1.19)
        (5, 0.96)        
    };

    \addplot[color=burgundy,style=solid,mark=diamond*,mark options={scale=1.1,solid}] plot coordinates {
        (2, 56.75)
        (3, 68.37)
        (4, 62.11)
        (5, 69.23)        
    };

    \addplot[color=navyblue,style=solid,mark=square*,mark options={scale=1.1,solid}] plot coordinates {
        (2, 20.13)
        (3, 18.35)
        (4, 15.91)
        (5, 11.54)        
    };

    \addplot[color=dartmouthgreen,style=solid,mark=*,mark options={scale=1.1,solid}] plot coordinates {
        (2, 3.11)
        (3, 2.92)
        (4, 1.9 )
        (5, 1.92)        
    };

    \legend{\acc~This work,\acc~Moses p.,\acc~Moses f.,\cov~This work,\cov~Moses p.,\cov~Moses f.}

    \end{axis}
    \node[] () [below = 1.6cm, right=0.3cm] at (ambplot.south east) {};
    \node[] () [below = 1.6cm, left=0.9cm] at (ambplot.south west) {};
\end{tikzpicture}
}
\subfloat[Evaluation of highly ambiguous compounds.]{%
  \usebox{\tempbox}
  \label{fig:amb_evalplot}
}\quad\subfloat[Evaluation of all compounds and highly ambiguous compounds only.]{\vbox to \ht\tempbox{%
\vfil
\hbox to 0.6\textwidth{
  \small
  \centering
  \begin{tabularx}{0.57\textwidth}{ l@{} c YY c YY c YY }
  \toprule
                     && \multicolumn{2}{@{}c@{}}{This work} && \multicolumn{2}{@{}c@{}}{Moses (partial)} && \multicolumn{2}{@{}c@{}}{Moses (full)} \\
                     \cmidrule{3-4} \cmidrule{6-7} \cmidrule{9-10} 
  Scenario           && \acc  & \cov && \acc  & \cov && \acc & \cov \\
  \toprule
  Full test set       && 27.43 & 58.45 && 18.04 & 31.41 && 6.57 & 13.75 \\
  \midrule
  2 splits           && 24.94 & 56.75&& 13.13 & 20.13&& 1.79 & 3.11 \\
  3 splits           && 21.10 & 68.37&& 8.04 & 18.35&& 1.21  & 2.92 \\
  4 splits           && 22.09 & 62.11&& 9.98  & 15.91&& 1.19 & 1.90  \\
  5 splits           && 24.04 & 69.23&& 9.62  & 11.54&& 0.96 & 1.92 \\
  \bottomrule
  \end{tabularx}
  \label{fig:amb_eval}
  \vspace{0.2cm}
 }
 \vfil
 }
}
\afterdoublefloat
\caption{Gold standard evaluation of compound splitting.}
\afterfig
\end{table}
Table~\ref{fig:amb_eval} presents accuracy and coverage for the compounds
within the different ambiguity scenarios.
To better visualize the trends for highly ambiguous compounds, we 
plot the accuracy and coverage scores in relation to the ambiguity of the
compounds in Table~\ref{fig:amb_evalplot}.
The analogy-based method outperforms the frequency-based baselines in both
coverage and accuracy.
While for the Moses splitter, the coverage decreases with increasing ambiguity, 
the opposite behavior is shown by our approach, as having more possible splits
results in a higher number of direction vectors increasing the likelihood of 
obtaining meaning-preserving splits. 
This experiment shows that the analogy-based compound splitter is advantageous for 
words that can potentially be explained by several candidate splits.

\section{Compound splitting for machine translation}
\label{sec:evaluation}

\paragraph{Translation setup}

We use the Moses decoder \cite{koehn2007moses} to train a phrase-based MT system 
on the English--German \emph{Common crawl} parallel corpus and \emph{WMT news test} 2010 (tuning).
Word alignment is performed with Giza++ \cite{och03:asc}. 
We use a 3rd order language model estimated using IRSTLM \cite{federico2008irstlm}, 
as well as lexicalized reordering.
The test data set is \emph{WMT news test} 2015,\footnote{\label{wmtnote2}\url{http://www.statmt.org/wmt15/translation-task.html}} 
which contains approx.\ 2100 de-en sentence pairs and 10000 tokens (with one reference translation).
We compare our method against a baseline translation system with no compound splitter, 
and the same system implementing Moses' default compound splitting tool.
The test set contains 2111 out-of-vocabulary word types (natural OOV words), 
which yields a total of 2765 unknown tokens, consisting mostly of compounds, 
brand names, and city names.
This implies that 22.16$\%$ (word types) resp.\ 7.15$\%$ (tokens) of the test 
corpus are unknown to the baseline system.


\paragraph{Translation experiments}

\newcommand{\siga}{\textsuperscript{\fontsize{5}{4}\selectfont A}}
\newcommand{\sigb}{\textsuperscript{\fontsize{5}{4}\selectfont B}}
\newcommand{\sigc}{\textsuperscript{\fontsize{5}{4}\selectfont C}}

\newcommand{\nosiga}{\phantom{\siga}}
\newcommand{\nosigb}{\phantom{\sigb}}
\newcommand{\nosigc}{\phantom{\sigc}}

\newcommand{\mincountwm}{\text{c(w)}}
\newcommand{\mincountw}{$\mincountwm$}

\begin{table}
\centering

\newcommand{\SPLTS}{\scriptsize Splits}
\newcommand{\BL}{\scriptsize BLEU}
\newcommand{\MTR}{\scriptsize MTR}

\newcommand{\win}[1]{\textbf{#1}}

\footnotesize
\begin{tabularx}{\textwidth}{l l YYY l YYY l YYY l YYY}
 \toprule
               && \multicolumn{3}{@{}c@{}}{(a) No comp.\ splitting}                                       &  & \multicolumn{3}{@{}c@{}}{(b) OOV only} &  & \multicolumn{3}{@{}c@{}}{(c) Rare: $\mincountwm < 20$} & & \multicolumn{3}{@{}c@{}}{(d) All words} \\ 
               \cmidrule{3-5} \cmidrule{7-9} \cmidrule{11-13} \cmidrule{15-17} 
               && \SPLTS   & \BL  & \MTR &  & \SPLTS  & \BL  & \MTR  &  & \SPLTS & \BL            & \MTR   &  & \SPLTS  & \BL        & \MTR \\ 
 \toprule
Moses splitter && \multirow{2}{*}{0}        & \multirow{2}{*}{17.6} &  \multirow{2}{*}{25.5}     &  & 226     & 17.6 & 25.7\siga  &  & 231    & 17.6           & 25.7                                   & & 244     & 17.9       & 25.8\siga        \\
This work      &&                           &                           &                        &  & 317     & 17.6 & 25.8\siga  &  & 744    & \win{18.2}\siga\sigb\sigc & \win{26.1}\siga\sigb\sigc   & & 1616    & 17.7       & 26.3\siga   \\
 \bottomrule
\end{tabularx}

\vspace{0.2cm}
\scriptsize
{\siga}\hspace{0.02cm}Stat.\ sign.\ against (a) at $p < 0.05$  \ \ 
{\sigb}\hspace{0.02cm}Stat.\ sign.\ against Moses splitter at same $\mincountwm$ at $p < 0.05$ \ \ 
{\sigc}\hspace{0.02cm}Stat.\ sign.\ against best Moses splitter (d) at $p < 0.05$  
\vspace{-0.1cm}
\caption{Translation results for various integration methods.}
\label{tbl:mtresults}
\afterfig
\end{table}

To test the analogy-based compound splitter on a realistic setting, 
we perform a standard machine translation task.
We translate a German text using 
 a translation baseline system with no compound handling (a), 
 a translation system integrating the standard Moses compound splitter tool trained using the best-performing settings,  
and a translation system using our analogy-based compound splitter.
We test the following basic methods of integration: 
Splitting only words that are OOV to the translation model (b),
splitting all words that occur less than 20 times 
in the training corpus (c),
and applying the compound splitters to every word in the datasets (d).
Table~\ref{tbl:mtresults} shows the results of these translation experiments. 
For each experiment, we report BLEU \cite{PapineniBLEU}, METEOR \cite{denkowski-lavie}, 
and the number of compound splits performed on the test set.
Statistical significance tests are performed using bootstrap resampling 
\cite{koehn2004statistical}.

\paragraph{Discussion}
The results show that when applied without restrictions, our method splits a large
number of words and leads to minor improvements.
When applied only to rare words 
the splitter produces statistically significant improvements in both BLEU and METEOR over 
the best frequency-based compound splitter.
This difference indicates that a better method for deciding which words the 
splitter should be applied to could lead to further improvements.
Overall, the output of the analogy-based compound splitter is more
beneficial to the machine translation system than the baseline splitter.


\section{Conclusion}
In this paper, we have studied whether regularities in the semantic word 
embedding space can be exploited to model the composition of compound words
based on analogy.
To approach this question, we made the following contributions:
First, we evaluated whether properties of compounds can be found in the
semantic vector space. 
We found that this space lends itself to modeling compounds based on their 
semantic head.
Based on this finding, we discussed how to extract compound transformations and
prototypes following the method of \newcite{soricut2015unsupervised} and proposed
an algorithm for applying these structures to compound splitting.
Our experiments show that the analogy-based compound splitter outperforms a
commonly used compound splitter on a gold standard task.
Our novel compound splitter is particularly adept at splitting 
highly ambiguous compounds.
Finally, we applied the analogy-based compound splitter in a machine translation
task and found that it compares favorably to the commonly used
shallow frequency-based method.

\paragraph{Acknowledgements}

Joachim Daiber is supported by the EXPERT (EXPloiting Empirical appRoaches 
to Translation) Initial Training Network (ITN) of the European Union's 
Seventh Framework Programme.
Stella Frank is supported by funding from the European Union’s Horizon
2020 research and innovation programme under grant agreement
Nr.~645452.

\bibliographystyle{acl}
\bibliography{sources}

\begin{thebibliography}{}

\bibitem[\protect\citename{Cap \bgroup et al.\egroup }2014]{Cap:2014}
Fabienne Cap, Alexander Fraser, Marion Weller, and Aoife Cahill.
\newblock 2014.
\newblock How to produce unseen teddy bears: Improved morphological processing
  of compounds in {SMT}.
\newblock In {\em Proceedings of the 14th Conference of the European Chapter of
  the Association for Computational Linguistics (EACL)}.

\bibitem[\protect\citename{Denkowski and Lavie}2014]{denkowski-lavie}
Michael Denkowski and Alon Lavie.
\newblock 2014.
\newblock Meteor universal: Language specific translation evaluation for any
  target language.
\newblock In {\em Proceedings of the Ninth Workshop on Statistical Machine
  Translation}.

\bibitem[\protect\citename{Federico \bgroup et al.\egroup
  }2008]{federico2008irstlm}
Marcello Federico, Nicola Bertoldi, and Mauro Cettolo.
\newblock 2008.
\newblock {IRSTLM}: An open source toolkit for handling large scale language
  models.
\newblock In {\em Proceedings of Interspeech 2008 - 9th Annual Conference of
  the International Speech Communication Association}.

\bibitem[\protect\citename{Fraser \bgroup et al.\egroup }2012]{Fraser:2012}
Alexander Fraser, Marion Weller, Aoife Cahill, and Fabienne Cap.
\newblock 2012.
\newblock Modeling inflection and word-formation in {SMT}.
\newblock In {\em Proceedings of the 13th Conference of the European Chapter of
  the Association for Computational Linguistics (EACL)}.

\bibitem[\protect\citename{Fritzinger and Fraser}2010]{Fritzinger:2010}
Fabienne Fritzinger and Alexander Fraser.
\newblock 2010.
\newblock How to avoid burning ducks: Combining linguistic analysis and corpus
  statistics for {German} compound processing.
\newblock In {\em Proceedings of the ACL 2010 Joint Fifth Workshop on
  Statistical Machine Translation and Metrics (MATR)}.

\bibitem[\protect\citename{Henrich and Hinrichs}2011]{henrich2011determining}
Verena Henrich and Erhard~W. Hinrichs.
\newblock 2011.
\newblock Determining immediate constituents of compounds in {GermaNet}.
\newblock In {\em Proceedings of the International Conference on Recent
  Advances in Natural Language Processing 2011}.

\bibitem[\protect\citename{Koehn and Knight}2003]{koehn2003empirical}
Philipp Koehn and Kevin Knight.
\newblock 2003.
\newblock Empirical methods for compound splitting.
\newblock In {\em Proceedings of the 10th Conference of the European Chapter of
  the Association for Computational Linguistics (EACL)}.

\bibitem[\protect\citename{Koehn \bgroup et al.\egroup }2007]{koehn2007moses}
Philipp Koehn, Hieu Hoang, Alexandra Birch, Chris Callison-Burch, Marcello
  Federico, Nicola Bertoldi, Brooke Cowan, Wade Shen, Christine Moran, Richard
  Zens, Chris Dyer, Ondrej Bojar, Alexandra Constantin, and Evan Herbst.
\newblock 2007.
\newblock Moses: Open source toolkit for statistical machine translation.
\newblock In {\em Proceedings of the 45th Annual Meeting of the Association for
  Computational Linguistics (ACL)}.

\bibitem[\protect\citename{Koehn}2004]{koehn2004statistical}
Philipp Koehn.
\newblock 2004.
\newblock Statistical significance tests for machine translation evaluation.
\newblock In {\em Proceedings of the 9th Conference on Empirical Methods in
  Natural Language Processing (EMNLP)}.

\bibitem[\protect\citename{Lieber and {\v S}tekauer}2009]{lieber2009oxford}
Rochelle Lieber and Pavol {\v S}tekauer.
\newblock 2009.
\newblock {\em The Oxford handbook of compounding}.
\newblock Oxford University Press.

\bibitem[\protect\citename{Mikolov \bgroup et al.\egroup
  }2013]{mikolov2013linguistic}
Tomas Mikolov, Wen-tau Yih, and Geoffrey Zweig.
\newblock 2013.
\newblock Linguistic regularities in continuous space word representations.
\newblock In {\em Proceedings of the 2013 Conference of the North American
  Chapter of the Association for Computational Linguistics (NAACL)}.

\bibitem[\protect\citename{Nie{\ss}en and Ney}2000]{Niessen:2000}
Sonja Nie{\ss}en and Hermann Ney.
\newblock 2000.
\newblock Improving {SMT} quality with morpho-syntactic analysis.
\newblock In {\em Proceedings of the 18th International Conference on
  Computational Linguistics (COLING)}.

\bibitem[\protect\citename{Och and Ney}2003]{och03:asc}
Franz~Josef Och and Hermann Ney.
\newblock 2003.
\newblock A systematic comparison of various statistical alignment models.
\newblock {\em Computational Linguistics}, 29(1):19--51.

\bibitem[\protect\citename{Papineni \bgroup et al.\egroup }2002]{PapineniBLEU}
Kishore Papineni, Salim Roukos, Todd Ward, and Wei-Jing Zhu.
\newblock 2002.
\newblock {BLEU}: A method for automatic evaluation of machine translation.
\newblock In {\em Proceedings of the 40th Annual Meeting of the Association for
  Computational Linguistics (ACL)}.

\bibitem[\protect\citename{Popovi{\'c} \bgroup et al.\egroup
  }2006]{Popovic:2006}
Maja Popovi{\'c}, Daniel Stein, and Hermann Ney.
\newblock 2006.
\newblock Statistical machine translation of {German} compound words.
\newblock In {\em Proceedings of FinTal - 5th International Conference on
  Natural Language Processing}.

\bibitem[\protect\citename{Reddy \bgroup et al.\egroup }2011]{Reddy:2011}
Siva Reddy, Diana McCarthy, and Suresh Manandhar.
\newblock 2011.
\newblock An empirical study on compositionality in compound nouns.
\newblock In {\em Proceedings of the 5th International Joint Conference on
  Natural Language Processing}.

\bibitem[\protect\citename{{Schulte im Walde} \bgroup et al.\egroup
  }2013]{SchulteimWalde:2013}
Sabine {Schulte im Walde}, Stefan M{\"u}ller, and Stephen Roller.
\newblock 2013.
\newblock Exploring vector space models to predict the compositionality of
  {German} noun-noun compounds.
\newblock In {\em Proceedings of the 2nd Joint Conference on Lexical and
  Computational Semantics (*SEM)}.

\bibitem[\protect\citename{Soricut and Och}2015]{soricut2015unsupervised}
Radu Soricut and Franz Och.
\newblock 2015.
\newblock Unsupervised morphology induction using word embeddings.
\newblock In {\em Proceedings of the 2015 Conference of the North American
  Chapter of the Association for Computational Linguistics (NAACL)}.

\bibitem[\protect\citename{Weller \bgroup et al.\egroup }2014]{Weller:2014}
Marion Weller, Fabienne Cap, Stefan M{\"u}ller, Sabine {Schulte im Walde}, and
  Alexander Fraser.
\newblock 2014.
\newblock Distinguishing degrees of compositionality in compound splitting for
  statistical machine translation.
\newblock In {\em Proceedings of the First Workshop on Computational Approaches
  to Compound Analysis (ComaComa) at COLING}.

\end{thebibliography}

\end{document}